\documentclass[
]{ceurart}

\begin{document}

\copyrightyear{2021}
\copyrightclause{Copyright for this paper by its authors.
  Use permitted under Creative Commons License Attribution 4.0
  International (CC BY 4.0).}

\conference{FIRE 2021: Forum for Information Retrieval Evaluation, December 13-17, 2021, India}
\title{Pegasus@Dravidian-CodeMix-HASOC2021: Analyzing Social Media Content for Detection of Offensive Text}

\author[1]{Pawan Kalyan Jada}[%
email=pawankj19c@iiitt.ac.in,
]

\author[1]{Konthala Yasaswini}[%
email=konthalay18c@iiitt.ac.in,
]

\author[1]{Karthik Puranik}[%
email=karthikp18c@iiitt.ac.in,
]

\address[1]{Indian Institute of Information Technology Tiruchirappalli}

\address[2]{Kongu Engineering College, Erode, Tamil Nadu, India}
\author[2]{Anbukkarasi Sampath}
[email=anbu.1318@gmail.com]
\address[3]{Sri Krishna Adithya College of Arts and Science, Coimbatore}
\author[3]{Sathiyaraj Thangasamy}
[email=sathiyarajt@skacas.ac.in]
\address[4]{Sultan Idris Education University, Tanjong Malim, Perak, Malaysia}

\author[4]{Kingston Pal Thamburaj}
[email=fkingston@gmail.com]

\begin{abstract}
To tackle the conundrum of detecting offensive comments/posts which are considerably informal, unstructured, miswritten and code-mixed, we introduce two inventive methods in this research paper. Offensive comments/posts on the social media platforms, can affect an individual, a group or underage alike. In order to classify comments/posts in two popular Dravidian languages, Tamil and Malayalam, as a part of the HASOC - DravidianCodeMix FIRE 2021 shared task, we employ two Transformer-based prototypes which successfully stood in the top 8 for all the tasks. The codes for our approach can be viewed and utilized\footnote{\url{https://github.com/PawanKalyanJada/hasoc}}.
\end{abstract}


\maketitle

\section{Introduction}
The term “Social media” provides a channel through which people engage in interactive communities and networks by creating, sharing, and exchanging thoughts and information. The growth rate of social media, especially Facebook and Twitter, has been exceptionally high since 2006. It has received users from almost all generations and all around the world. Users can interact and connect with others and form communities through social media.  It allows users to share their ideas, views and information openly on various topics. This gives license to the users to write hateful and offensive comments sometimes. People come from a variety of racial backgrounds and hold a diversity of belief systems. This often causes for a conflict of opinions during their interactions on social media platforms. Many derogatory content target individuals based on their skin colour, gender, caste, nationality, religion, race, ethnicity. Due to the COVID-19 pandemic, the internet community has become more popular than it has ever been \citep{9418446}. The amount of false narratives and derogatory remarks shared on online platforms has shot through the roof. A large number of social media users share malicious posts despite understanding that they are infringing on their right to free expression. This may have a negative and adverse impact on user's mental health.  Various social media platforms restrict and minimize the profane comments by employing new rules and techniques.

Online hate speech and offensive content produces challenges to the society. The detection of hate speech is quite a daunting task, as the precise understanding of the speech largely depends on the circumstances it is being used in. Due to the extreme enormity of the internet and the growing number of online users, as well as the obscurity of the users, manually detecting and removing hate speech and profane information is a time-consuming and challenging job.

Code-mixing often entails the use of two languages to produce a third language that incorporates aspects from each in a functionally comprehensible manner. Low-resource languages such as Tamil and Malayalam are gaining significant attention alongside English on social networking platforms. The majority of the data on social media for these under-resourced languages is code-mixed. Tamil is a Dravidian language spoken by Tamils in India and Sri Lanka, as well as the Tamil community worldwide \cite{hande2021domain}. The official recognition of the language is in India, Sri Lanka, and Singapore. Tamil was the first to be classified as a classical language of India and is one of the longest-surviving classical languages in the world. Tamil has the oldest extant literature among Dravidian languages \cite{chakravarthi-etal-2020-corpus}. Malayalam is a Dravidian language spoken in southern India, having official language status in the Indian state of Kerala as well as the Union Territories of Lakshadweep and Puducherry. Malayalam scripts are alpha-syllabic, a type of “Abugida” writing system that is partially alphabetic and partly syllable-based.

This paper presents our work for the shared task on offensive language detection of code-mixed text in Dravidian languages (Malayalam-English and Tamil-English) at HASOC - DravidianCodeMix FIRE 2021. The rest of the paper is summarized as follows,
\ref{Section 2} presents a discussion on the previous works on Offensive
Language Detection in Dravidian Languages. \ref{Section 3} entails a detailed task description and analysis of the datasets
for Tamil, Malayalam, and Kannada. In \ref{Section 4}
we present a description of the models used for the tasks.

\section{Related work}
\label{Section 2}
There has been a tremendous advancement in the research of offensive language detection over the past few years. On social media, hate speech in the form of racist and sexist statements is quite commonplace. In  \citeauthor{waseem-hovy-2016-hateful}, the authors provided a dataset of 16k tweets annotated for hate speech and analysed the features that help detect hate speech in the corpus. The authors of \citeauthor{davidson2017automated} used logistical regression to extract N-gram TF-IDF features from tweets and categorize each tweet into hate, offensive, and non-offensive categories. For identifying abusive language, the authors of \citeauthor{hassan-etal-2020-alt} experimented with Support Vector Machines (SVMs) \citep{708428} trained on character and word-level features, Deep Neural Networks (DNNs) and Bidirectional Encoder Representations from Transformers (BERT) \citep{devlin2019bert}. An Ensembling based approach which is based on hybridization of Naive Bayes, SVM, Linear Regression, and SGD classifiers was developed and tested on a Hindi-English code-mix dataset which outperformed the state-of-the-art systems and baseline models \cite{9342241}.

In \citeauthor{liu2019nuli}, the authors experimented with various classifiers which includes linear model with features of word unigrams, word2vec, and Hatebase; word-based Long Short-Term Memory (LSTM) \citep{articlelstm}; fine-tuned
Bidirectional Encoder Representation from Transformer (BERT). \citeauthor{hande-etal-2020-kancmd} created Kannada CodeMixed Dataset (KanCMD), a multitask learning dataset for sentiment analysis and offensive language identification. We work with several transformer-based models to classify social media comments as hope speech or not hope speech in English, Malayalam and Tamil languages. Various transformer-based models were fine-tuned to classify social media comments in English, Malayalam and Tamil languages into hope speech and non-hope speech labels \cite{puranik2021iiittltedieacl2021hope, chakravarthi-2020-hopeedi, chakravarthi-muralidaran-2021-findings}. In \citeauthor{yasaswini-etal-2021-iiitt}, the authors developed a model, CNN-BiLSTM, which has a layer of 1D convolutional layer followed by a dropout layer and then a bidirectional LSTM
layer for identifying offensive language comments which are often code-mixed. The authors of \citeauthor{hande2021hope} introduced a Dual-Channel
BERT4Hope approach employed by fine-tuning a
language model based on BERT on the code-mixed
data and its translation in English. Soft-voting is implemented on the fine-tuned transformer models to determine if any sentence contains information
about an event that has occurred or not \cite{kalyan-etal-2021-iiitt}.

\section{Task description and dataset}
\label{Section 3}
The main aim of the HASOC Shared task is to identify offensive content in the code-mixed comments/posts in the Dravidian languages collected
from social media \cite{HASOC-dravidiancodemix-2021}. There are two tasks, in which task 1 is a message-level label classification task, systems have to classify a given YouTube comment in Tamil into offensive or not-offensive. Task 2 is also a message-level label classification task, in which systems have to classify a given tweet in code mixed Tamil and Malayalam into offensive or not-offensive.

We are provided with two different datasets for the two subtasks. The training dataset provided for task 1 comprised of 5877 YouTube comments in Tamil classified into offensive or not-offensive. The task 1 was limited to Tamil language. The dataset is also observed to be imbalanced. The training dataset provided for task 2 comprised of 4000 tweets in code mixed Tamil and 3999 tweets in code mixed Malayalam. The training and validation datasets provided for task 2 are well-balanced.   

\begin{table}[htbp]
    \centering 
    \begin{tabular}{|l|r|r|r|} \hline
       \textbf{Language}  &  \textbf{Tamil (Task 1)} & \textbf{Tamil (Task 2)} & \textbf{Malayalam (Task 2)} \\ \hline  
       not offensive  & 4,724 & 2,020 & 2047 \\
       offensive  & 1,153 & 1,980 & 1952\\ \hline
       Total & 5,877 & 4,000 & 3,999\\ \hline
    \end{tabular}
    \caption{Class-wise distribution of the training set for both the Tasks}
    \label{train_set}
\end{table}

\begin{figure*}[htbp]
\centering
\includegraphics[width=13cm,height=9.0cm]{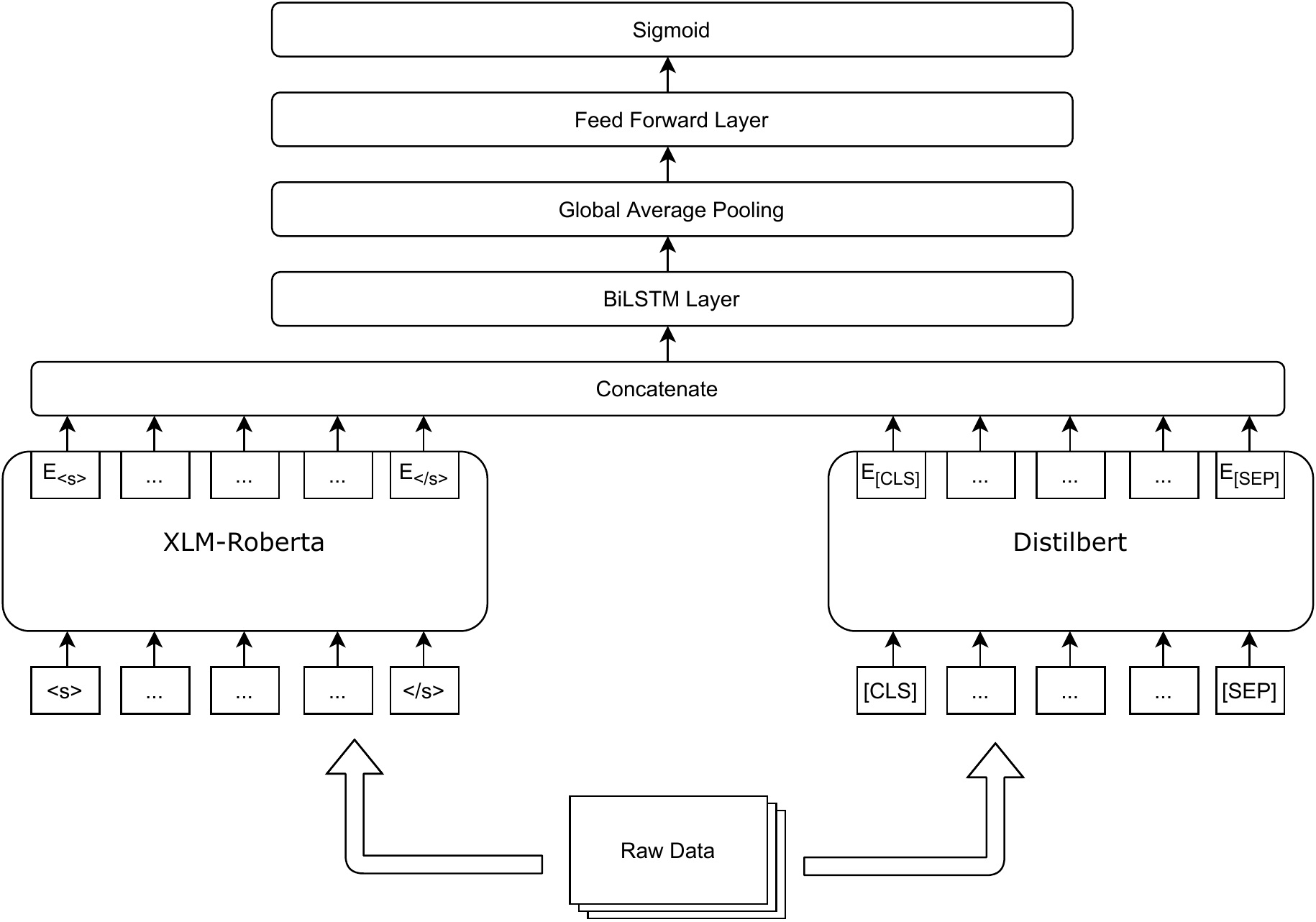}
\caption{Model Architecture employed for Task 1}
\label{fig1}
\end{figure*}

\section{System Description}
\label{Section 4}
\subsection{Task 1}
We fine tune transformer based language models for this task. Firstly, emoticons and flags were cleaned from the dataset. Then the sentences were converted to lower case as some of the samples contains English text between them. We then pass the sequences through two pre-trained models namely XLM-R and DistilBERT extracting the embeddings from both. These embeddings were then concatenated before being passed through the BiLSTM layers \cite{xu2019sentiment}, eventually being pooled on a global average scale. These are fed to some Fully Connected layers and an Activation Function of sigmoid to get the probability scores as shown in Figure \ref{fig1}. By concatenating the embeddings, we expect that the model we created can benefit from knowledge of both the NLP models employed for the task, helping to distinguish better among the classes.

\subsection{Task 2}
\begin{figure*}[htbp]
\centering
\includegraphics[width=13cm,height=9.5cm]{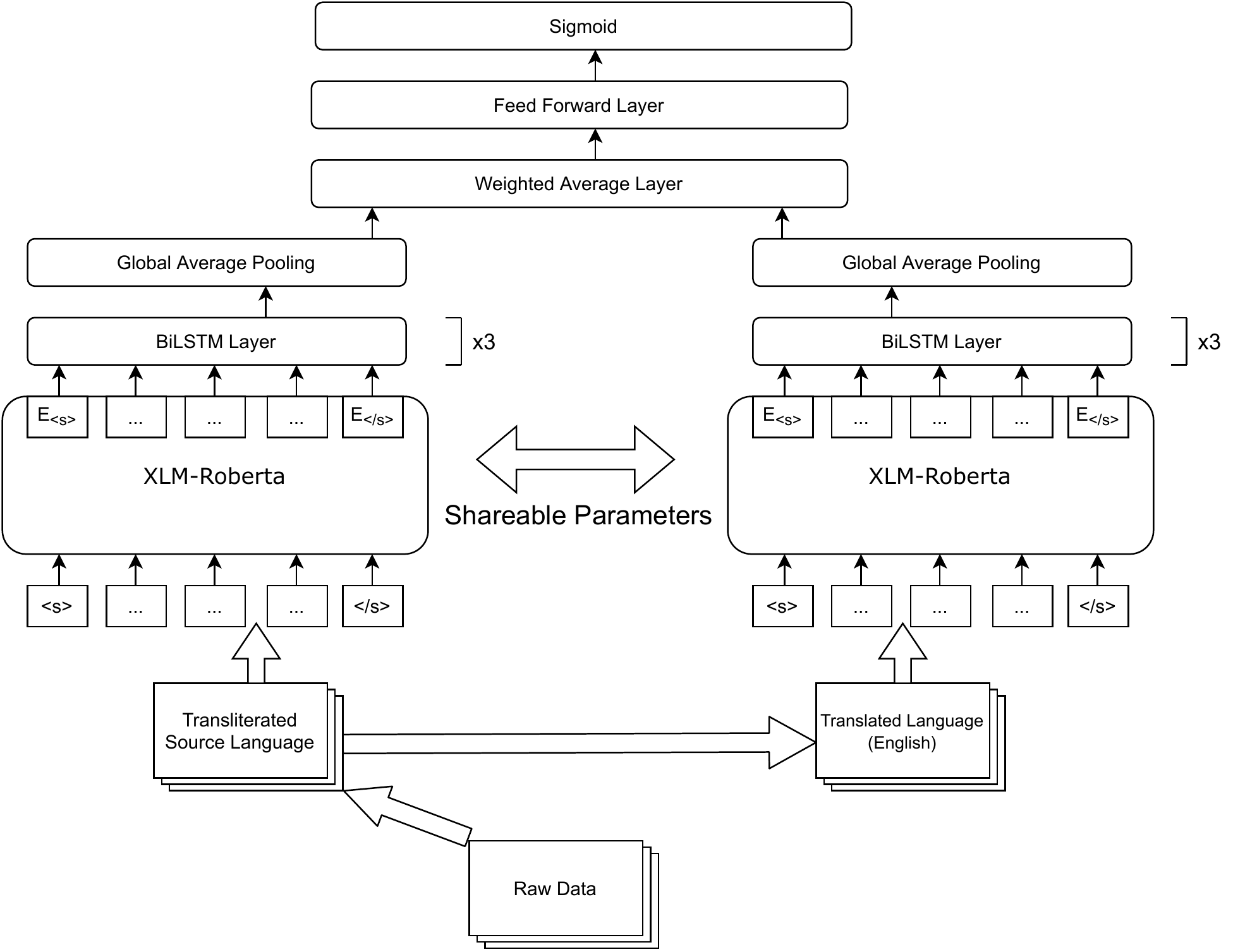}
\caption{Model Architecture employed for Task 2}
\label{fig2}
\end{figure*}

For this task, we first transliterate \cite{regmi2010understanding} the Tamil sentences library in the English script to the native Tamil script by usage of “indic-transliteration” library\footnote{\url{https://pypi.org/project/indic-transliteration/}}. We then translate these sentences to English using the Google Translate API\cite{wu2016googles}. We then clean this parallel corpora of sentences by removing punctuations, stripping unwanted spaces at the end and converting the English sentences to a lower case. After the preprocessing, we tokenize these sequences using a tokenizer of a multilingual model, XLM-R. These tokens of Tamil and English are fed through the same XLM-R model and then passed through BiLSTM layers and a pooling layer at the end. Then we compute the weighted average of the Tamil and English vectors, with weights of 0.7 for Tamil and 0.3 for English. An Activation Function of sigmoid \cite{yin2003flexible} is also applied at the end, deriving the probability scores required to classify a sentence. The entire architecture is shown in Figure \ref{fig2}. The same technique is done for Malayalam sentences as well, here the sentences being transliterated to Malayalam and the weights being 0.6 for Malayalam and 0.4 for English. These weights were set upon experimentation with various values and then selecting the best from all of them. Refer table \ref{parameters} for parameters used in this task.

\begin{table}[htbp]
\centering
    \begin{tabular}{|lr|} \hline
       \textbf{Parameters}  &  \textbf{Values} \\ \hline  
       Number of LSTM layers  & 3 \\
       Number of LSTM units  & 128 \\
       Batch Size & 32 \\ 
       Max Length & 128\\
       Optimizer & Adam \\ 
       Learning Rate & 1e-3 \\ 
       Activation Function & Sigmoid \\ 
       Loss Function & cross-entropy \\  \hline
    \end{tabular}
    \caption{Parameters used for training the model in Task 2}
    \label{parameters}
\end{table}

\section{Methodology}
\subsection{XLM-RoBERTa}
XLM-R \cite{conneau2019unsupervised} is a multilingual language model that achieved state-of-the-art results in all the multiple cross lingual benchmarks. One of the reason for the unparalleled performance is that it was trained on a mammoth 2.5 TB of CommonCrawl data \cite{smith2013dirt}. It was trained with MLM loss as it's objective on 100 different languages, and it shares similar training routine as the one employed for RoBERTa \cite{liu2019roberta} which is the reason the model is called XLM RoBERTa. XLM Roberta is fine-tuned for both of these tasks in a different architecture employed for the specific task. For this task we use \textit{\textbf{xlm-roberta-base}} which consists of 12-layers, 768-hidden-state, 8-heads and a parameter size of 270M. The reason for selecting XLM-R is it outperformed other models in all the multilingual benchmarks.

\subsection{DistilBERT}
DistilBERT \cite{sanh2019distilbert} is BERT's distilled version. It employs a triple loss language model that combines language modelling, distillation and cosine-distance losses. The two distillation losses in the triple loss have a significant influence on model performance. We fine tune \textit{\textbf{distilbert-base-multilingual-cased}}, which is distilled from the mBERT checkpoint, for our cause in Task 1. It is known to have 40\% less number of parameters than mBERT and runs 60\% faster than it. The model consists of 6 layers, 768 dimensions, and 12 Attention heads, with a total of around 134 million parameters. We chose DistilBERT due to the less size of the model making it extract the word embeddings quicker.

\section{Results}

\subsection{Task 1}

The system description model on the Tamil language for this task gave a promising F1 score of 0.810. Embeddings from two of the most efficient multilingual pretrained models, XLM-R and DistilBERT were concatenated to extract their significant features. Further, the use of BiLSTM layers improves the accuracy as the information being fed doubles. The BiLSTM layers contains two LSTM's which take the input from the forward and backward directions and, thus, enhancing the context. However, one of the drawbacks which causes an impediment is the class imbalance between the “offensive” and “not-offensive” sentences in the test dataset. Also, XLM-R and DistilBERT follow BERT-based architectures and hence, the embeddings produced don't generate huge variations. Employment of pretrained models belonging to other architectures could produce higher accuracies. Increasing the number of models also might enhance the quality of embeddings produced, and thus, boosting the F1 scores further to 0.810.

\begin{table*}[!h]
    \centering
    \begin{tabular*}{\tblwidth}{@{} L|RRR|RRR|RRRR@{}}
    \toprule
    \multicolumn{10}{c}{\textbf{Task 1}}\\
    \midrule
    &\multicolumn{3}{c}{\textbf{Not offensive}}&\multicolumn{3}{c}{\textbf{Offensive}} & \multicolumn{4}{c}{\textbf{Overall}}\\
    \midrule
    \textbf{Language} & \textbf{P} & \textbf{R} & \textbf{F}& \textbf{P} & \textbf{R} & \textbf{F}& \textbf{P} & \textbf{R} & \textbf{F} & \textbf{Acc}\\
    \midrule
    \textbf{Tamil}& 0.888  &   0.875   &  0.882 &0.468    & 0.500    & 0.484&0.812   &  0.807    & 0.810 & 0.807\\
    \toprule
    \multicolumn{10}{c}{\textbf{Task 2}}\\
    \midrule
    &\multicolumn{3}{c}{\textbf{Not offensive}}&\multicolumn{3}{c}{\textbf{Offensive}}&\multicolumn{4}{c}{\textbf{Overall}}\\
    \midrule
    \textbf{Language} & \textbf{P} & \textbf{R} & \textbf{F}& \textbf{P} & \textbf{R} & \textbf{F}& \textbf{P} & \textbf{R} & \textbf{F}& \textbf{Acc}\\
    \midrule
    \textbf{Tamil}& 0.657   &  0.865    & 0.746& 0.596    & 0.306 &    0.405 & 0.633   &  0.644    & 0.612 & 0.644\\
    \midrule
    \textbf{Malayalam}& 0.817    & 0.640 &    0.718& 0.483    & 0.701    & 0.572&0.708 &    0.660 &    0.670 & 0.660\\
    \bottomrule
    \end{tabular*}
    \caption{Weighted F1 scores for task 1 and 2 by our system model on the test dataset where, P=Precision, R=Recall, F=F1-score, Acc=Accuracy}
    \label{tab:my_label}
\end{table*}

\begin{table*}[]
    \centering
    \begin{tabular*}{\tblwidth}{@{} L|RRR|RRR|RRRR@{}}
    \toprule
    \multicolumn{10}{c}{\textbf{Task 1}}\\
    \midrule
    &\multicolumn{3}{c}{\textbf{Not offensive}}&\multicolumn{3}{c}{\textbf{Offensive}} & \multicolumn{4}{c}{\textbf{Overall}}\\
    \midrule
    \textbf{Language} & \textbf{P} & \textbf{R} & \textbf{F}& \textbf{P} & \textbf{R} & \textbf{F}& \textbf{P} & \textbf{R} & \textbf{F} & \textbf{Acc}\\
    \midrule
    \textbf{Tamil}& 0.868  &   0.891    & 0.879 &0.500     &0.446    & 0.471&0.796  &   0.804   &  0.799& 0.804\\
    \toprule
    \multicolumn{10}{c}{\textbf{Task 2}}\\
    \midrule
    &\multicolumn{3}{c}{\textbf{Not offensive}}&\multicolumn{3}{c}{\textbf{Offensive}}&\multicolumn{4}{c}{\textbf{Overall}}\\
    \midrule
    \textbf{Language} & \textbf{P} & \textbf{R} & \textbf{F}& \textbf{P} & \textbf{R} & \textbf{F}& \textbf{P} & \textbf{R} & \textbf{F}& \textbf{Acc}\\
    \midrule
    \textbf{Tamil}& 0.865  &   0.957  &   0.909& 0.953     &0.855    & 0.901& 0.910   &  0.905  &   0.905& 0.905\\
    \midrule
    \textbf{Malayalam}& 0.758   &  0.729&     0.744& 0.742    & 0.770 &    0.756& 0.750   &  0.750  &   0.750& 0.750\\
    \bottomrule
    \end{tabular*}
    \caption{Weighted F1 scores for task 1 and 2 by our system model on the validation dataset where, P=Precision, R=Recall, F=F1-score, Acc=Accuracy}
    \label{val}
\end{table*}

\subsection{Task 2}
It was found out that our model gave an F1 score of 0.612 for Tamil and 0.670 for Malayalam. The dataset contains Tamil and Malayalam comments written in the Roman script, which is hard for the multilingual pretrained models trained on the native scripts to comprehend. Transliterating these sentences to the native language can prove to increase the F1 scores. Furthermore, we know that the models like XLM-R is trained on a large corpus of English sentences. Thus, the English translations of these transliterated dataset plays a huge role in further fine-tuning of the model. With the test dataset again containing sentences in native languages written in the Roman script, it was essential to give a higher precedence to the transliterated tokens over the translated tokens \cite{Puranik2021AttentiveFO}. BiLSTM once again plays its role in ameliorating the results by increasing the information being fed. 

However, we can never be definite of the accuracy in the transliterations and translations. Reduced quality if these sentences can affect the fine-tuning of the model significantly, and hence, lowering the F1 scores. Class imbalance prevails in this dataset too. With the ratio of not offensive to offensive in the range of 2:1, the model seems to find it arduous to predict the offensive sentences efficiently. It is observed that the F1 scores between the not offensive and offensive differ by 0.15 to 0.3. This also impacts the overall F1 score for this task. However, as we can see in Table \ref{val}, the difference between the F1-scores of the “offensive” and “not-offensive” labels on the validation dataset didn't seem to vary much. Table \ref{tab:my_label} tabulates the detailed weighted F1 scores for the test dataset.

\section{Error Analysis}

Few of the notable sentences where we felt that the sentences were misclassified have been discussed in this section. In task 1, 528 sentences were classified correctly in Tamil, while 126 failed to be classified well. There can be several reasons for this misclassification. Sometimes, the presence of offensive words doesn't ensure that the sentence is “Offensive” and vice-versa. Comments are also filled with sarcasms, puns and typographical errors which have high probability of getting classified wrongly. Second task in Tamil has 645 sentences classified correctly and 356 wrongly. In Malayalam, 713 are correct and 238 wrong. We have discussed few of the Tamil sentences,\\
\newline
Task 1: \\
\textbf{adey kirukka nalla paru,,,google unaku theriyuma,,,, 2rs eppadi ellam pesura,,,Sanghis}\\
This sentence is tagged as “not offensive”, but it is directed towards North Indians and probably as a reply to another comment/post.\\
\newline
Task 2:\\
\textbf{Inta treylar kuta parkkira matiri illai.. Itai tiyettar la poy parkkanuma}\\
The sentence is classified as offensive, but it is just a review which states, “this trailer itself isn't good. Does someone have to go to the theatres too to watch this?”.\\

Another major drawback was the poor quality of translations by the Google API. The accuracy of classifications would have been better if the translations were of good quality in all the cases. For example,\\
\newline
\textbf{Sentence 1}: tl vere oru ss kandu\\
\textbf{\textit{There is no other way}} is a very good translation of the Malayalam sentence and the model is able to learn and predict well.\\
\newline
\textbf{Sentence 2}: aga surya um jothikaum etho plan pani taga pola,not,Aga Surya Uma Jyotika is like something Plan Bani Daka\\
\textbf{Aga Surya Um Jyotikaum Something like Plan snow} is an example of how some sentences get partially got converted to English.\\
\newline
\textbf{Sentence 3}: aaiiii jolly yellam onnah polam onnah polam oannaaa polam update app to view\\
\textbf{IEE Jolly Yellam Onnah Bolam Onnah Bolam Oannaa Polam Uptade App To Viev} is the supposed to be the English translation of the above sentence. We can see that the quality of the translation is very bad.

\section{Conclusion}

Offensive language detection in social media posts presents to be a significant task due to social and marketing rationale. For the task of offensive language detection in code-mixed Dravidian languages Tamil and Malayalam, we introduce our research in this paper. For task 1, we extract the embeddings from XLM-R and DistilBERT, we concatenate them and pass them through BiLSTM layers. This model managed to give an F1 score of 0.810. Similarly, for task 2 we transliterate the dataset in which the Dravidian language is written in the Roman script and then, translate them into English and fine-tune the XLM-R model on it. This model gives us F1 scores of 0.612 for Tamil and 0.670 for Malayalam. Neglecting the fact that the English translations were of a poor quality, the model achieves very decent F1 scores for both the languages and, thus, opening a gateway for more research in this field. 

\bibliography{sample-ceur}

\begin{thebibliography}{28}
\expandafter\ifx\csname natexlab\endcsname\relax\def\natexlab#1{#1}\fi
\providecommand{\url}[1]{\texttt{#1}}
\providecommand{\href}[2]{#2}
\providecommand{\path}[1]{#1}
\providecommand{\DOIprefix}{doi:}
\providecommand{\ArXivprefix}{arXiv:}
\providecommand{\URLprefix}{URL: }
\providecommand{\Pubmedprefix}{pmid:}
\providecommand{\doi}[1]{\href{http://dx.doi.org/#1}{\path{#1}}}
\providecommand{\Pubmed}[1]{\href{pmid:#1}{\path{#1}}}
\providecommand{\bibinfo}[2]{#2}
\ifx\xfnm\relax \def\xfnm[#1]{\unskip,\space#1}\fi
\bibitem[{Hande et~al.(2021{\natexlab{a}})Hande, Puranik, Priyadharshini,
  Thavareesan, and Chakravarthi}]{9418446}
\bibinfo{author}{A.~Hande}, \bibinfo{author}{K.~Puranik},
  \bibinfo{author}{R.~Priyadharshini}, \bibinfo{author}{S.~Thavareesan},
  \bibinfo{author}{B.~R. Chakravarthi},
\newblock \bibinfo{title}{Evaluating pretrained transformer-based models for
  covid-19 fake news detection},
\newblock in: \bibinfo{booktitle}{2021 5th International Conference on
  Computing Methodologies and Communication (ICCMC)},
  \bibinfo{year}{2021}{\natexlab{a}}, pp. \bibinfo{pages}{766--772}.
  \DOIprefix\doi{10.1109/ICCMC51019.2021.9418446}.
\bibitem[{Hande et~al.(2021{\natexlab{b}})Hande, Puranik, Priyadharshini, and
  Chakravarthi}]{hande2021domain}
\bibinfo{author}{A.~Hande}, \bibinfo{author}{K.~Puranik},
  \bibinfo{author}{R.~Priyadharshini}, \bibinfo{author}{B.~R. Chakravarthi},
\newblock \bibinfo{title}{Domain identification of scientific articles using
  transfer learning and ensembles},
\newblock in: \bibinfo{booktitle}{Trends and Applications in Knowledge
  Discovery and Data Mining: PAKDD 2021 Workshops, WSPA, MLMEIN, SDPRA, DARAI,
  and AI4EPT, Delhi, India, May 11, 2021 Proceedings 25},
  \bibinfo{organization}{Springer International Publishing},
  \bibinfo{year}{2021}{\natexlab{b}}, pp. \bibinfo{pages}{88--97}.
\bibitem[{Chakravarthi et~al.(2020)Chakravarthi, Muralidaran, Priyadharshini,
  and McCrae}]{chakravarthi-etal-2020-corpus}
\bibinfo{author}{B.~R. Chakravarthi}, \bibinfo{author}{V.~Muralidaran},
  \bibinfo{author}{R.~Priyadharshini}, \bibinfo{author}{J.~P. McCrae},
\newblock \bibinfo{title}{Corpus creation for sentiment analysis in code-mixed
  {T}amil-{E}nglish text},
\newblock in: \bibinfo{booktitle}{Proceedings of the 1st Joint Workshop on
  Spoken Language Technologies for Under-resourced languages (SLTU) and
  Collaboration and Computing for Under-Resourced Languages (CCURL)},
  \bibinfo{publisher}{European Language Resources association},
  \bibinfo{address}{Marseille, France}, \bibinfo{year}{2020}, pp.
  \bibinfo{pages}{202--210}. \URLprefix
  \url{https://aclanthology.org/2020.sltu-1.28}.
\bibitem[{Waseem and Hovy(2016)}]{waseem-hovy-2016-hateful}
\bibinfo{author}{Z.~Waseem}, \bibinfo{author}{D.~Hovy},
\newblock \bibinfo{title}{Hateful symbols or hateful people? predictive
  features for hate speech detection on {T}witter},
\newblock in: \bibinfo{booktitle}{Proceedings of the {NAACL} Student Research
  Workshop}, \bibinfo{publisher}{Association for Computational Linguistics},
  \bibinfo{address}{San Diego, California}, \bibinfo{year}{2016}, pp.
  \bibinfo{pages}{88--93}. \URLprefix \url{https://aclanthology.org/N16-2013}.
  \DOIprefix\doi{10.18653/v1/N16-2013}.
\bibitem[{Davidson et~al.(2017)Davidson, Warmsley, Macy, and
  Weber}]{davidson2017automated}
\bibinfo{author}{T.~Davidson}, \bibinfo{author}{D.~Warmsley},
  \bibinfo{author}{M.~Macy}, \bibinfo{author}{I.~Weber},
  \bibinfo{title}{Automated hate speech detection and the problem of offensive
  language}, \bibinfo{year}{2017}. \href{http://arxiv.org/abs/1703.04009}{{\tt
  arXiv:1703.04009}}.
\bibitem[{Hassan et~al.(2020)Hassan, Samih, Mubarak, Abdelali, Rashed, and
  Chowdhury}]{hassan-etal-2020-alt}
\bibinfo{author}{S.~Hassan}, \bibinfo{author}{Y.~Samih},
  \bibinfo{author}{H.~Mubarak}, \bibinfo{author}{A.~Abdelali},
  \bibinfo{author}{A.~Rashed}, \bibinfo{author}{S.~A. Chowdhury},
\newblock \bibinfo{title}{{ALT} submission for {OSACT} shared task on offensive
  language detection},
\newblock in: \bibinfo{booktitle}{Proceedings of the 4th Workshop on
  Open-Source Arabic Corpora and Processing Tools, with a Shared Task on
  Offensive Language Detection}, \bibinfo{publisher}{European Language Resource
  Association}, \bibinfo{address}{Marseille, France}, \bibinfo{year}{2020}, pp.
  \bibinfo{pages}{61--65}. \URLprefix
  \url{https://aclanthology.org/2020.osact-1.9}.
\bibitem[{Hearst et~al.(1998)Hearst, Dumais, Osuna, Platt, and
  Scholkopf}]{708428}
\bibinfo{author}{M.~Hearst}, \bibinfo{author}{S.~Dumais},
  \bibinfo{author}{E.~Osuna}, \bibinfo{author}{J.~Platt},
  \bibinfo{author}{B.~Scholkopf},
\newblock \bibinfo{title}{Support vector machines},
\newblock \bibinfo{journal}{IEEE Intelligent Systems and their Applications}
  \bibinfo{volume}{13} (\bibinfo{year}{1998}) \bibinfo{pages}{18--28}.
  \DOIprefix\doi{10.1109/5254.708428}.
\bibitem[{Devlin et~al.(2019)Devlin, Chang, Lee, and
  Toutanova}]{devlin2019bert}
\bibinfo{author}{J.~Devlin}, \bibinfo{author}{M.-W. Chang},
  \bibinfo{author}{K.~Lee}, \bibinfo{author}{K.~Toutanova},
  \bibinfo{title}{Bert: Pre-training of deep bidirectional transformers for
  language understanding}, \bibinfo{year}{2019}.
  \href{http://arxiv.org/abs/1810.04805}{{\tt arXiv:1810.04805}}.
\bibitem[{Yadav et~al.(2020)Yadav, Lamba, Gupta, Gupta, Karmakar, and
  Saini}]{9342241}
\bibinfo{author}{K.~Yadav}, \bibinfo{author}{A.~Lamba},
  \bibinfo{author}{D.~Gupta}, \bibinfo{author}{A.~Gupta},
  \bibinfo{author}{P.~Karmakar}, \bibinfo{author}{S.~Saini},
\newblock \bibinfo{title}{Bi-lstm and ensemble based bilingual sentiment
  analysis for a code-mixed hindi-english social media text},
\newblock in: \bibinfo{booktitle}{2020 IEEE 17th India Council International
  Conference (INDICON)}, \bibinfo{year}{2020}, pp. \bibinfo{pages}{1--6}.
  \DOIprefix\doi{10.1109/INDICON49873.2020.9342241}.
\bibitem[{Liu et~al.(2019)Liu, Li, and Zou}]{liu2019nuli}
\bibinfo{author}{P.~Liu}, \bibinfo{author}{W.~Li}, \bibinfo{author}{L.~Zou},
\newblock \bibinfo{title}{Nuli at semeval-2019 task 6: Transfer learning for
  offensive language detection using bidirectional transformers},
\newblock in: \bibinfo{booktitle}{Proceedings of the 13th international
  workshop on semantic evaluation}, \bibinfo{year}{2019}, pp.
  \bibinfo{pages}{87--91}.
\bibitem[{Hochreiter and Schmidhuber(1997)}]{articlelstm}
\bibinfo{author}{S.~Hochreiter}, \bibinfo{author}{J.~Schmidhuber},
\newblock \bibinfo{title}{Long short-term memory},
\newblock \bibinfo{journal}{Neural computation} \bibinfo{volume}{9}
  (\bibinfo{year}{1997}) \bibinfo{pages}{1735--80}.
  \DOIprefix\doi{10.1162/neco.1997.9.8.1735}.
\bibitem[{Hande et~al.(2020)Hande, Priyadharshini, and
  Chakravarthi}]{hande-etal-2020-kancmd}
\bibinfo{author}{A.~Hande}, \bibinfo{author}{R.~Priyadharshini},
  \bibinfo{author}{B.~R. Chakravarthi},
\newblock \bibinfo{title}{{K}an{CMD}: {K}annada {C}ode{M}ixed dataset for
  sentiment analysis and offensive language detection},
\newblock in: \bibinfo{booktitle}{Proceedings of the Third Workshop on
  Computational Modeling of People's Opinions, Personality, and Emotion's in
  Social Media}, \bibinfo{publisher}{Association for Computational
  Linguistics}, \bibinfo{address}{Barcelona, Spain (Online)},
  \bibinfo{year}{2020}, pp. \bibinfo{pages}{54--63}. \URLprefix
  \url{https://aclanthology.org/2020.peoples-1.6}.
\bibitem[{Puranik et~al.(2021)Puranik, Hande, Priyadharshini, Thavareesan, and
  Chakravarthi}]{puranik2021iiittltedieacl2021hope}
\bibinfo{author}{K.~Puranik}, \bibinfo{author}{A.~Hande},
  \bibinfo{author}{R.~Priyadharshini}, \bibinfo{author}{S.~Thavareesan},
  \bibinfo{author}{B.~R. Chakravarthi},
  \bibinfo{title}{Iiitt@lt-edi-eacl2021-hope speech detection: There is always
  hope in transformers}, \bibinfo{year}{2021}.
  \href{http://arxiv.org/abs/2104.09066}{{\tt arXiv:2104.09066}}.
\bibitem[{Chakravarthi(2020)}]{chakravarthi-2020-hopeedi}
\bibinfo{author}{B.~R. Chakravarthi},
\newblock \bibinfo{title}{{H}ope{EDI}: A multilingual hope speech detection
  dataset for equality, diversity, and inclusion},
\newblock in: \bibinfo{booktitle}{Proceedings of the Third Workshop on
  Computational Modeling of People's Opinions, Personality, and Emotion's in
  Social Media}, \bibinfo{publisher}{Association for Computational
  Linguistics}, \bibinfo{address}{Barcelona, Spain (Online)},
  \bibinfo{year}{2020}, pp. \bibinfo{pages}{41--53}. \URLprefix
  \url{https://aclanthology.org/2020.peoples-1.5}.
\bibitem[{Chakravarthi and
  Muralidaran(2021)}]{chakravarthi-muralidaran-2021-findings}
\bibinfo{author}{B.~R. Chakravarthi}, \bibinfo{author}{V.~Muralidaran},
\newblock \bibinfo{title}{Findings of the shared task on hope speech detection
  for equality, diversity, and inclusion},
\newblock in: \bibinfo{booktitle}{Proceedings of the First Workshop on Language
  Technology for Equality, Diversity and Inclusion},
  \bibinfo{publisher}{Association for Computational Linguistics},
  \bibinfo{address}{Kyiv}, \bibinfo{year}{2021}, pp. \bibinfo{pages}{61--72}.
  \URLprefix \url{https://aclanthology.org/2021.ltedi-1.8}.
\bibitem[{Yasaswini et~al.(2021)Yasaswini, Puranik, Hande, Priyadharshini,
  Thavareesan, and Chakravarthi}]{yasaswini-etal-2021-iiitt}
\bibinfo{author}{K.~Yasaswini}, \bibinfo{author}{K.~Puranik},
  \bibinfo{author}{A.~Hande}, \bibinfo{author}{R.~Priyadharshini},
  \bibinfo{author}{S.~Thavareesan}, \bibinfo{author}{B.~R. Chakravarthi},
\newblock \bibinfo{title}{{IIITT}@{D}ravidian{L}ang{T}ech-{EACL}2021: Transfer
  learning for offensive language detection in {D}ravidian languages},
\newblock in: \bibinfo{booktitle}{Proceedings of the First Workshop on Speech
  and Language Technologies for Dravidian Languages},
  \bibinfo{publisher}{Association for Computational Linguistics},
  \bibinfo{address}{Kyiv}, \bibinfo{year}{2021}, pp. \bibinfo{pages}{187--194}.
  \URLprefix \url{https://aclanthology.org/2021.dravidianlangtech-1.25}.
\bibitem[{Hande et~al.(2021)Hande, Priyadharshini, Sampath, Thamburaj,
  Chandran, and Chakravarthi}]{hande2021hope}
\bibinfo{author}{A.~Hande}, \bibinfo{author}{R.~Priyadharshini},
  \bibinfo{author}{A.~Sampath}, \bibinfo{author}{K.~P. Thamburaj},
  \bibinfo{author}{P.~Chandran}, \bibinfo{author}{B.~R. Chakravarthi},
  \bibinfo{title}{Hope speech detection in under-resourced kannada language},
  \bibinfo{year}{2021}. \href{http://arxiv.org/abs/2108.04616}{{\tt
  arXiv:2108.04616}}.
\bibitem[{Kalyan et~al.(2021)Kalyan, Reddy, Hande, Priyadharshini, Sakuntharaj,
  and Chakravarthi}]{kalyan-etal-2021-iiitt}
\bibinfo{author}{P.~Kalyan}, \bibinfo{author}{D.~Reddy},
  \bibinfo{author}{A.~Hande}, \bibinfo{author}{R.~Priyadharshini},
  \bibinfo{author}{R.~Sakuntharaj}, \bibinfo{author}{B.~R. Chakravarthi},
\newblock \bibinfo{title}{{IIITT} at {CASE} 2021 task 1: Leveraging pretrained
  language models for multilingual protest detection},
\newblock in: \bibinfo{booktitle}{Proceedings of the 4th Workshop on Challenges
  and Applications of Automated Extraction of Socio-political Events from Text
  (CASE 2021)}, \bibinfo{publisher}{Association for Computational Linguistics},
  \bibinfo{address}{Online}, \bibinfo{year}{2021}, pp.
  \bibinfo{pages}{98--104}. \URLprefix
  \url{https://aclanthology.org/2021.case-1.13}.
  \DOIprefix\doi{10.18653/v1/2021.case-1.13}.
\bibitem[{Chakravarthi et~al.(2021)Chakravarthi, Kumaresan, Sakuntharaj,
  Madasamy, Thavareesan, B, Chinnaudayar~Navaneethakrishnan, McCrae, and
  Mandl}]{HASOC-dravidiancodemix-2021}
\bibinfo{author}{B.~R. Chakravarthi}, \bibinfo{author}{P.~K. Kumaresan},
  \bibinfo{author}{R.~Sakuntharaj}, \bibinfo{author}{A.~K. Madasamy},
  \bibinfo{author}{S.~Thavareesan}, \bibinfo{author}{P.~B},
  \bibinfo{author}{S.~Chinnaudayar~Navaneethakrishnan}, \bibinfo{author}{J.~P.
  McCrae}, \bibinfo{author}{T.~Mandl},
\newblock \bibinfo{title}{{Overview of the HASOC-DravidianCodeMix Shared Task
  on Offensive Language Detection in Tamil and Malayalam}},
\newblock in: \bibinfo{booktitle}{Working Notes of FIRE 2021 - Forum for
  Information Retrieval Evaluation}, \bibinfo{publisher}{CEUR},
  \bibinfo{year}{2021}.
\bibitem[{Xu et~al.(2019)Xu, Meng, Qiu, Yu, and Wu}]{xu2019sentiment}
\bibinfo{author}{G.~Xu}, \bibinfo{author}{Y.~Meng}, \bibinfo{author}{X.~Qiu},
  \bibinfo{author}{Z.~Yu}, \bibinfo{author}{X.~Wu},
\newblock \bibinfo{title}{Sentiment analysis of comment texts based on bilstm},
\newblock \bibinfo{journal}{Ieee Access} \bibinfo{volume}{7}
  (\bibinfo{year}{2019}) \bibinfo{pages}{51522--51532}.
\bibitem[{Regmi et~al.(2010)Regmi, Naidoo, and
  Pilkington}]{regmi2010understanding}
\bibinfo{author}{K.~Regmi}, \bibinfo{author}{J.~Naidoo},
  \bibinfo{author}{P.~Pilkington},
\newblock \bibinfo{title}{Understanding the processes of translation and
  transliteration in qualitative research},
\newblock \bibinfo{journal}{International Journal of Qualitative Methods}
  \bibinfo{volume}{9} (\bibinfo{year}{2010}) \bibinfo{pages}{16--26}.
\bibitem[{Wu et~al.(2016)Wu, Schuster, Chen, Le, Norouzi, Macherey, Krikun,
  Cao, Gao, Macherey, Klingner, Shah, Johnson, Liu, Łukasz Kaiser, Gouws,
  Kato, Kudo, Kazawa, Stevens, Kurian, Patil, Wang, Young, Smith, Riesa,
  Rudnick, Vinyals, Corrado, Hughes, and Dean}]{wu2016googles}
\bibinfo{author}{Y.~Wu}, \bibinfo{author}{M.~Schuster},
  \bibinfo{author}{Z.~Chen}, \bibinfo{author}{Q.~V. Le},
  \bibinfo{author}{M.~Norouzi}, \bibinfo{author}{W.~Macherey},
  \bibinfo{author}{M.~Krikun}, \bibinfo{author}{Y.~Cao},
  \bibinfo{author}{Q.~Gao}, \bibinfo{author}{K.~Macherey},
  \bibinfo{author}{J.~Klingner}, \bibinfo{author}{A.~Shah},
  \bibinfo{author}{M.~Johnson}, \bibinfo{author}{X.~Liu},
  \bibinfo{author}{Łukasz Kaiser}, \bibinfo{author}{S.~Gouws},
  \bibinfo{author}{Y.~Kato}, \bibinfo{author}{T.~Kudo},
  \bibinfo{author}{H.~Kazawa}, \bibinfo{author}{K.~Stevens},
  \bibinfo{author}{G.~Kurian}, \bibinfo{author}{N.~Patil},
  \bibinfo{author}{W.~Wang}, \bibinfo{author}{C.~Young},
  \bibinfo{author}{J.~Smith}, \bibinfo{author}{J.~Riesa},
  \bibinfo{author}{A.~Rudnick}, \bibinfo{author}{O.~Vinyals},
  \bibinfo{author}{G.~Corrado}, \bibinfo{author}{M.~Hughes},
  \bibinfo{author}{J.~Dean}, \bibinfo{title}{Google's neural machine
  translation system: Bridging the gap between human and machine translation},
  \bibinfo{year}{2016}. \href{http://arxiv.org/abs/1609.08144}{{\tt
  arXiv:1609.08144}}.
\bibitem[{Yin et~al.(2003)Yin, Goudriaan, Lantinga, Vos, and
  Spiertz}]{yin2003flexible}
\bibinfo{author}{X.~Yin}, \bibinfo{author}{J.~Goudriaan},
  \bibinfo{author}{E.~A. Lantinga}, \bibinfo{author}{J.~Vos},
  \bibinfo{author}{H.~J. Spiertz},
\newblock \bibinfo{title}{A flexible sigmoid function of determinate growth},
\newblock \bibinfo{journal}{Annals of botany} \bibinfo{volume}{91}
  (\bibinfo{year}{2003}) \bibinfo{pages}{361--371}.
\bibitem[{Conneau et~al.(2019)Conneau, Khandelwal, Goyal, Chaudhary, Wenzek,
  Guzm{\'a}n, Grave, Ott, Zettlemoyer, and Stoyanov}]{conneau2019unsupervised}
\bibinfo{author}{A.~Conneau}, \bibinfo{author}{K.~Khandelwal},
  \bibinfo{author}{N.~Goyal}, \bibinfo{author}{V.~Chaudhary},
  \bibinfo{author}{G.~Wenzek}, \bibinfo{author}{F.~Guzm{\'a}n},
  \bibinfo{author}{E.~Grave}, \bibinfo{author}{M.~Ott},
  \bibinfo{author}{L.~Zettlemoyer}, \bibinfo{author}{V.~Stoyanov},
\newblock \bibinfo{title}{Unsupervised cross-lingual representation learning at
  scale},
\newblock \bibinfo{journal}{arXiv preprint arXiv:1911.02116}
  (\bibinfo{year}{2019}).
\bibitem[{Smith et~al.(2013)Smith, Saint-Amand, Plamad{\u{a}}, Koehn,
  Callison-Burch, and Lopez}]{smith2013dirt}
\bibinfo{author}{J.~Smith}, \bibinfo{author}{H.~Saint-Amand},
  \bibinfo{author}{M.~Plamad{\u{a}}}, \bibinfo{author}{P.~Koehn},
  \bibinfo{author}{C.~Callison-Burch}, \bibinfo{author}{A.~Lopez},
\newblock \bibinfo{title}{Dirt cheap web-scale parallel text from the common
  crawl},
\newblock in: \bibinfo{booktitle}{Proceedings of the 51st Annual Meeting of the
  Association for Computational Linguistics (Volume 1: Long Papers)},
  \bibinfo{year}{2013}, pp. \bibinfo{pages}{1374--1383}.
\bibitem[{Liu et~al.(2019)Liu, Ott, Goyal, Du, Joshi, Chen, Levy, Lewis,
  Zettlemoyer, and Stoyanov}]{liu2019roberta}
\bibinfo{author}{Y.~Liu}, \bibinfo{author}{M.~Ott}, \bibinfo{author}{N.~Goyal},
  \bibinfo{author}{J.~Du}, \bibinfo{author}{M.~Joshi},
  \bibinfo{author}{D.~Chen}, \bibinfo{author}{O.~Levy},
  \bibinfo{author}{M.~Lewis}, \bibinfo{author}{L.~Zettlemoyer},
  \bibinfo{author}{V.~Stoyanov},
\newblock \bibinfo{title}{Roberta: A robustly optimized bert pretraining
  approach},
\newblock \bibinfo{journal}{arXiv preprint arXiv:1907.11692}
  (\bibinfo{year}{2019}).
\bibitem[{Sanh et~al.(2019)Sanh, Debut, Chaumond, and
  Wolf}]{sanh2019distilbert}
\bibinfo{author}{V.~Sanh}, \bibinfo{author}{L.~Debut},
  \bibinfo{author}{J.~Chaumond}, \bibinfo{author}{T.~Wolf},
\newblock \bibinfo{title}{Distilbert, a distilled version of bert: smaller,
  faster, cheaper and lighter},
\newblock \bibinfo{journal}{arXiv preprint arXiv:1910.01108}
  (\bibinfo{year}{2019}).
\bibitem[{Puranik et~al.(2021)Puranik, Hande, Priyadharshini, Durairaj,
  Sampath, Thamburaj, and Chakravarthi}]{Puranik2021AttentiveFO}
\bibinfo{author}{K.~Puranik}, \bibinfo{author}{A.~Hande},
  \bibinfo{author}{R.~Priyadharshini}, \bibinfo{author}{T.~Durairaj},
  \bibinfo{author}{A.~Sampath}, \bibinfo{author}{K.~Thamburaj},
  \bibinfo{author}{B.~R. Chakravarthi},
\newblock \bibinfo{title}{Attentive fine-tuning of transformers for translation
  of low-resourced languages @loresmt 2021},
\newblock \bibinfo{year}{2021}.

\end{thebibliography}
\appendix

\end{document}